\documentclass[letterpaper, 10 pt, conference]{ieeeconf}  

\IEEEoverridecommandlockouts 

\usepackage[nolist,nohyperlinks]{acronym}
\usepackage{makecell}
\usepackage{wasysym}
\usepackage{amssymb}
\usepackage{lmodern}
\usepackage{arydshln}
\usepackage{graphicx}

\begin{acronym}
    \acro{NEEM}{Narrative-Enabled Episodic Memory}
    \acrodefplural{NEEM}{Narrative-Enabled Episodic Memories}
    \acro{DUL}{DOLCE+DnS-Ultralite}
    \acro{SOMA}{Socio-physical Model of Activities}
    \acro{IRT}{Intersensory Redundancy Theory}
    \acro{PRAC}{Probabilistic Action Cores}
    \acro{LLM}{Large Language Model}
    \acrodefplural{LLM}{Large Language Models}
    \acro{DNN}{Deep Neural Network}
    \acro{LTM}{Long-Term Memory}
    \acro{STM}{Short-Term Memory}
    \acro{JSON-LD}{JSON Linked Data}
    \acro{eKG}{Episodic Knowledge Graph}
    \acro{RDF}{Resource Description Framework}
    \acro{VR}{Virtual Reality}
    \acro{SOMA}{Socio-Physical Model of Activities}
    \acro{AGI}{Artificial General Intelligence}
    \acro{ITL}{Interactive Task Learning}
    \acro{CCTL}{Co-constructive Task Learning} 
    \acro{HHI}{Human-Human Interaction}
    \acro{HRI}{Human-Robot Interaction} 
    \acro{GUI}{Graphical User Interface}
    \acro{UML}{Unified Modeling Language}
\end{acronym}

\title{\LARGE \bf Towards a cognitive architecture to enable natural language interaction in co-constructive task learning}

\author{Manuel Scheibl$^{1}$, Birte Richter$^{1}$, Alissa Müller$^{2}$, Michael Beetz$^{3}$, Britta Wrede$^{1}$%
\thanks{1 Medical Assistance Systems Group, Medical School OWL, University of Bielefeld, Germany}%
\thanks{2 Machine Learning Group, Technical Faculty, University of Bielefeld, Germany}
\thanks{3 AICOR, University of Bremen, Germany}
}

\begin{document}

\maketitle
\thispagestyle{empty}
\pagestyle{empty}

\begin{abstract}
This research addresses the question, which characteristics a cognitive architecture must have to leverage the benefits of natural language in \ac{CCTL}.
To provide context, we first discuss \ac{ITL}, the mechanisms of the human memory system, and the significance of natural language and multi-modality.  Next, we examine the current state of cognitive architectures, analyzing their capabilities to inform a concept of \ac{CCTL} grounded in multiple sources. 
We then integrate insights from various research domains to develop a unified framework. Finally, we conclude by identifying the remaining challenges and requirements necessary to achieve \ac{CCTL} in \ac{HRI}.
\end{abstract}


\section{Introduction}
\label{sec:introduction}
Learning new actions is one of the most relevant targets in robotics research. It has for a long time been treated as a uni-directional process in which the teacher demonstrates actions to the robot. The learning is focused on generalizing them for novel contexts \cite{ravichandar2020recent}. Thus, due to the complexity of representing and learning actions in terms of goals, sub-goals and generalizable movements, the focus has been on how to achieve an optimal movement. 
However, it has also been noted that humans learn actions more efficiently by interaction. This has lead to the proposal of a new way of teaching: \acf{ITL} \cite{laird_cognitive_2010}. The fundamental idea behind \ac{ITL} lies in the fact that interaction is a bi-directional process that involves action demonstration and explanation from the teacher but also trials of action execution by the learner, which allows the teacher to directly react to errors or misconceptions \cite{wrede2009towards}. It has been shown that humans have an intuitive propensity and capability to teach and scaffold, and that children have an innate tendency to attend to and make use of such supporting means \cite{csibra2006social}.
Wrede et al. therefore suggest that, in order to learn in an interactive way robots need: (1) multi-modal integration at all processing levels (2) derivation of top-down strategies from bottom-up processes, and (3) integration of single modules into an interactive system in order to facilitate the first two desiderata \cite{wrede2009towards}.

\begin{figure}
    \centering
    \includegraphics[width=0.9\linewidth]{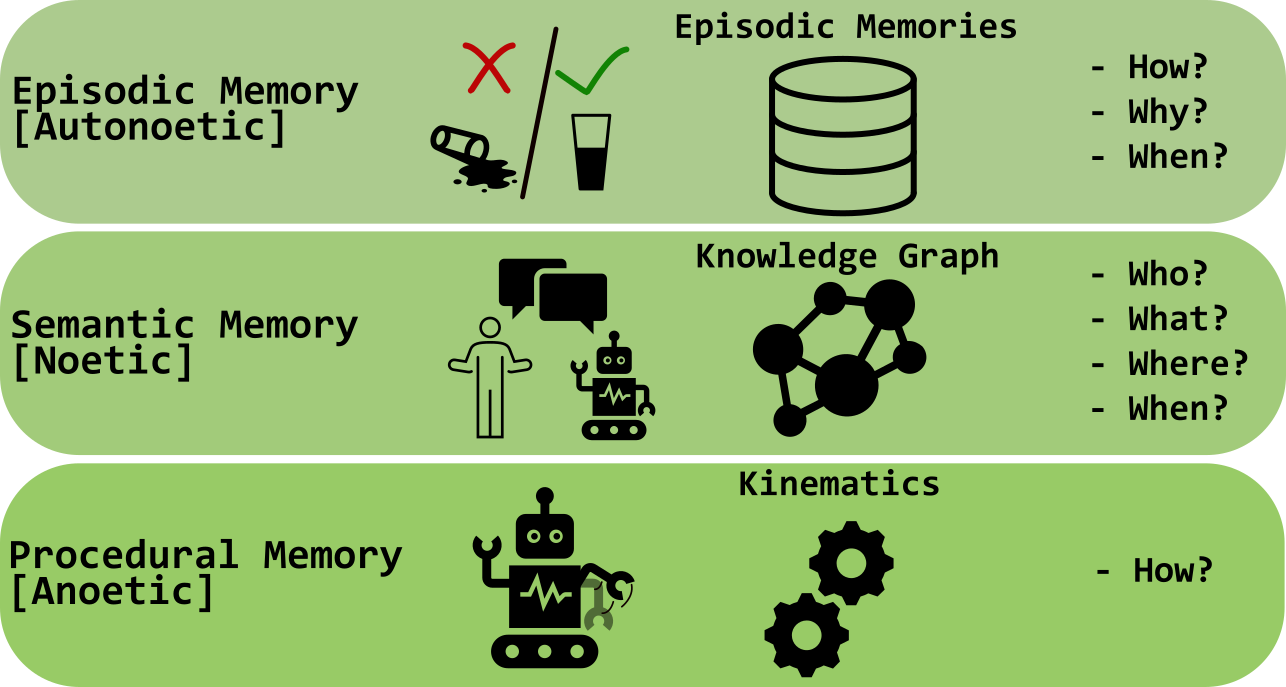}
    \caption{The memory systems in the robotic realm with the associated level of consciousness.}
    \label{fig:memory_subsystems}
\end{figure}

More recently, these suggestions have been taken up in \cite{thomaz2019interaction} where the problem is conceptualized as an issue of bi-directionally transferring information between the learner and the human via a shared world which allows to ground objects and actions. Additionally, four methods for a teacher to guide the learner are proposed: (1) through directing attention, (2) providing back-channels as a feedback means, (3) Providing a task description through verbal communication using natural language or some description language, and (4) demonstrating a task to the learner \cite{thomaz2019interaction} . This requires that the robot learner is capable of interpreting these scaffolding devices and to integrate the content to its existing knowledge base.
In the following this paper will, thus, address the following research questions:
\begin{itemize}
\item What characteristics must a cognitive architecture possess to enable \acf{CCTL} in human-robot interaction?
\item What roles does human speech play, and what features must a cognitive architecture support to fully leverage natural language?
\end{itemize}

To answer these questions, we explore the importance of multi-modality in robotic learning, with a focus on verbal interactions. 
Additionally, we analyze the key features of cognitive architectures and their relevance for \ac{ITL}.
Finally, we integrate interdisciplinary concepts — such as verbal input, episodic memory, pragmatic frames, acoustic packaging, and co-construction — into a unified framework to support \ac{CCTL}.

\section{Background}
\label{sec:background}
\subsection{Interactive and Co-Constructive Task Learning}
\label{sec:background:interactive_task_learning}
\textit{Co-Construction} is the process of creating a mutual understanding of the matter of concern during the interaction and implicitly optimizing communication to improve the learning process, and therefore extends the concept of \acf{ITL}~\cite{vollmer_interactive_2023}. \textit{Interaction} is defined as the mutual influence between people or social groups and the change it causes in behavior and attitude \cite{bergius_soziale_2022}.

\textit{Pragmatic Frames} are fundamental to learning. Introduced by Bruner for \ac{HHI}, they conceptualize the structure of learning processes and mark the content of an interaction that is relevant for learning success \cite{bruner_childs_1985}. In doing so, they can be used to guide the attention of a student to the most important part of an interaction. Vollmer et al. elevate this concept to be used in \ac{HRI} \cite{vollmer_pragmatic_2016}. In addition, by conceptualizing interactions, Pragmatic Frames can be used to compare and evaluate interaction strategies for teaching and learning.
Pragmatic Frames enable the robot to pay attention to the part of interaction which is relevant to acquire novel knowledge. Since episodic memories also contain irrelevant data for the acquisition of a new skill, the concept of \textit{Pragmatic Frames} is obligatory to effectively learn from prior memories.

\subsection{Memory}
\label{sec:background:memory}
The human memory system is central to the collective advancements of mankind and is the reason why humans are able to build on the achievements of previous generations to drive ideas and gather knowledge \cite{murray_evolutionary_2020}.
It consists of various subsystems with different specializations \cite{tulving_how_1985}. Firstly, memory can be divided into \ac{STM} and \ac{LTM} which can be divided further into procedural, semantic and episodic memory \cite{tulving_how_1985}.
The procedural memory stores knowledge about how to apply the learned skills. It is non-declarative and thus not actively accessible (anoetic).
Semantic memory stores generalizable declarative knowledge that is not sorted on a timescale.
In contrast, episodic memory keeps past experiences with a mental time stamp and in an autonoetic fashion, allowing the actor to mentally relive them \cite{tulving_episodic_2002}.
Since subsystems are interconnected, detailed episodic memories are only possible with deep semantic memory.
Vice versa, episodic memories are the source of knowledge that will eventually be stored in the semantic memory \cite{greenberg_interdependence_2010}.

The human memory is great at gathering knowledge to deduce novel insights and understand complex actions, motivating the transfer of the insights from neuroscience to the realm of robotics.
In Figure \ref{fig:memory_subsystems}, the subsystems of the \ac{LTM} are mapped to their representation in the robotic realm and to the questions that they give an answer to. The episodic memory primarily provides answers on how to do something. This knowledge can be leveraged to infer causal rules about why or when something is done.
The semantic memory is often represented as a knowledge graph of concepts and their interrelations. It gives answers on who or what to interact with, where to locate something, or when to do something. It is crucial for the understanding of a conversation.
The procedural memory is specific to the robot's body. It stores the kinematic procedures to, e.g. move the limbs.

\subsection{Natural Language}
\label{sec:background:natural_language}
Humans need to interact with the outside world by their five senses to gather knowledge, and more than 70\% of all processed information comes from sight \cite{wang_multi-modal_2024,rosenblum_see_2011}. 
Yet, natural language is the best way to quickly communicate thoughts from one individual to another, while also being efficient in terms of its informational density \cite{tomas_mikolov_roadmap_2018,cohen_role_1992} and used in scaffolding strategies during teaching scenarios. Movements, combined with speech, structure a task into partial actions that serve as guidance for the learner \cite{schillingmann_computational_2009}. This process, called \textit{Acoustic Packaging}, guides the attention towards the relevant pieces of an action \cite{wrede_making_2013} and it does also prove effective in robotics \cite{wang_multi-modal_2024,lohse_enabling_2013}.

\begin{figure}
    \centering
    \includegraphics[width=0.8\linewidth]{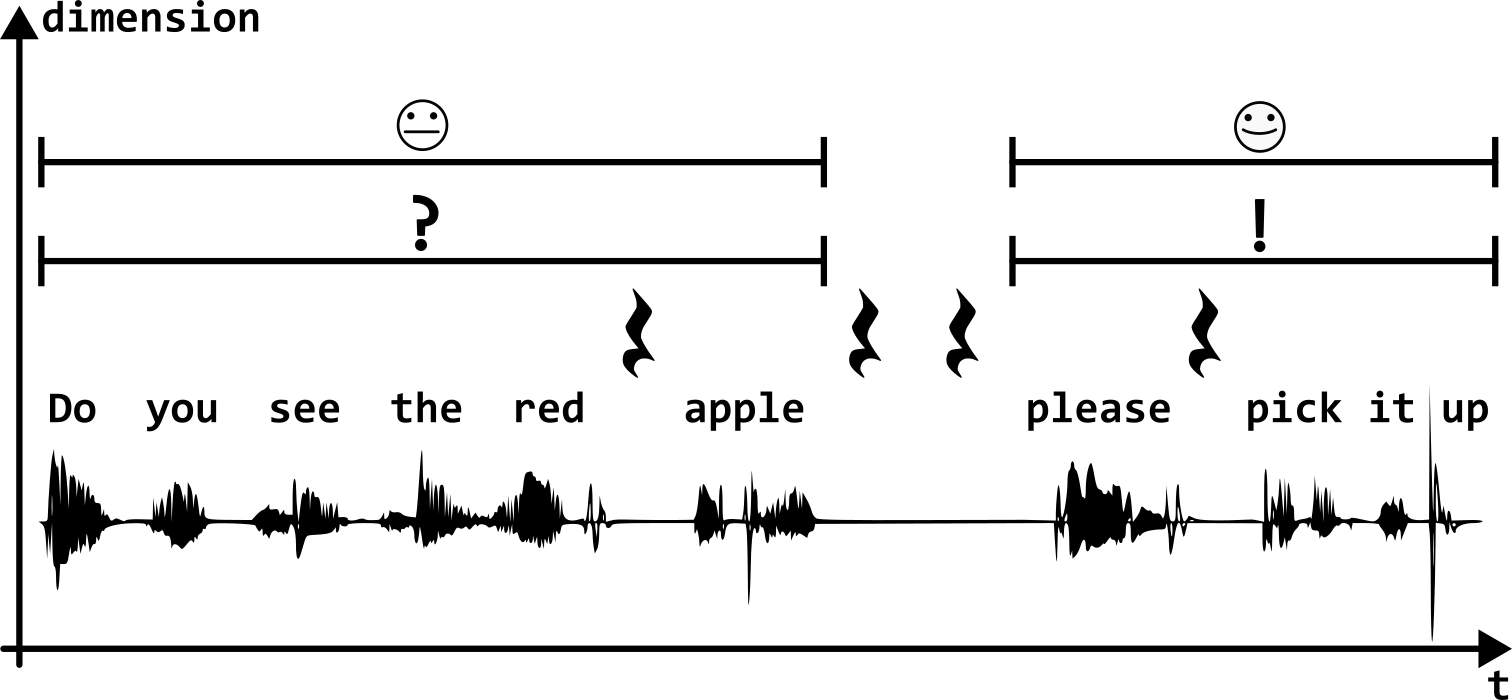}
    \caption{Raw audio can be differently tokenized. Words can be extracted, the speaker's mood can be analyzed, and it can also be assessed if a question was uttered. Moreover, the hesitation between words may convey information.}
    \label{fig:tokens_in_speech}
\end{figure}

To process input and acquire knowledge, the human brain needs to memorize information, e.g. utterances. However, it is only capable of memorizing verbal input as tokens that might be of higher dimensions to store different aspects, e.g., idiosyncratic features \cite{garner_processing_1975,mullennix_stimulus_1990,port_how_2007}. 
Pronunciation or pauses are also valuable to disambiguate the meaning of a word in case of \textit{heteronyms} and to understand the speaker's intention.
For humans, the tokenization depends on aspects e.g. auditory and linguistic experience and language \cite{port_how_2007,hawkins_polysp_2010,iverson_mapping_1995,logan_training_1991,strange_speech_1995,werker_cross-language_1984}.
Some discussed tokenizations are depicted in Figure \ref{fig:tokens_in_speech}. It indicates that there are many dimensions of information that natural language conveys simultaneously.
Depending on the requirements and purpose of a cognitive architecture, the applied tokenization varies. However, tokenizing the raw audio input does always come with an irretrievable loss of information.

\subsection{Multi-Modality}
\label{sec:background:multi-modality}
\begin{figure}
    \centering
    \includegraphics[width=0.8\linewidth]{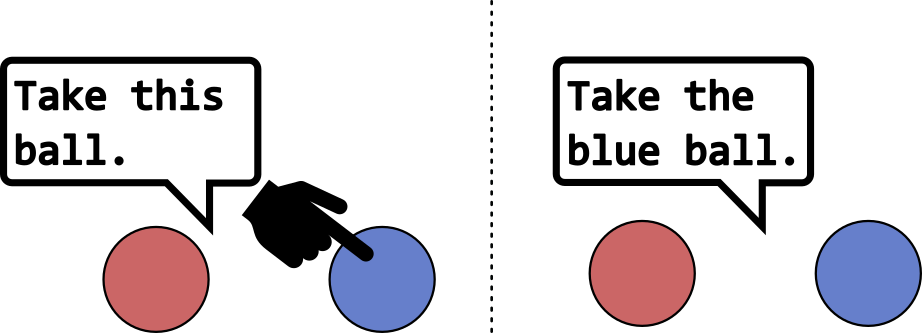}
    \caption{Multi-modality to define a task. In the left example, the task to pick a ball is only fully disclosed by using the verbal and visual modality. In the right example, the task is specified verbally.}
    \label{fig:input_multimodality}
\end{figure}

In \ac{HHI}, multiple modalities are often needed to disclose an interaction's purpose, as visualized in Figure \ref{fig:input_multimodality}. The left side shows that the target of an interaction might be named verbally, but an additional pointing action is needed to fully disclose the task whereas, on the right side, it is fully described verbally.

Each employed modality adds information to obtain new knowledge from. The theory of the \ac{IRT} states that multiple modalities support the understanding and memorization of an action and its meaning \cite{schillingmann_computational_2009}. Also, it is shown that multi-modality enhances perceptual learning \cite{lickliter_intersensory_2017} with verbal and visual input being the most significant modalities for effective learning \cite{magoon_effect_1977}. 
We expect the same benefits from multi-modality when applied to \ac{HRI}.

\section{Related Work}
\label{sec:related_work}
\subsection{Characteristics of Cognitive Architectures}
Little research is conducted on the requirements and characteristics for cognitive architectures to enable \ac{CCTL}.
Sun et al. formulate four top-level requirements for cognitive architectures.
The \textit{ecological realism} demands everyday activities to be executable. The \textit{cognitive realism} criteria demands the architecture to be inspired by cognitive science, without getting lost in the details. The \textit{bio-evolutionary realism} emphasizes that there should not be any overcomplicated mechanisms that would create a hard break between human- and animal-inspired architectures. 
Lastly, the \textit{eclecticism of methodologies} states that the technical realization of the architecture may vary among implementations and that different paradigms and approaches are eligible \cite{sun_desiderata_2004}.

In many ways similar to the goal of designing a human-like cognitive architecture for robotics, Laird et al. describe the environment, task, and agent characteristics that an \ac{AGI} should be built around and phrase twelve requirements that are mainly concerned with the representation and acquirement of knowledge \cite{laird_cognitive_2010}. These requirements are further elaborated in Section \ref{sec:concept:characteristics}.
Adams et al. enrich the requirements by a definition of competency areas that a human covers and which a cognitive architecture can be evaluated on \cite{adams_mapping_2012}.

Already, there is a variety of implemented cognitive architectures.
A major review has been conducted by Kotseruba et al. \cite{kotseruba_40_2020}. They classify the architectures by the categories symbolic (explicit knowledge), emergent (sub-symbolic), and hybrid (combining the two) and assess them by the supported input modalities, the memory architecture and by the  strategy for learning and reasoning. However, the review does not analyze the capabilities and usability for \ac{HRI}.
One recent approach to implementing the human memory system is done by Peller-Konrad et al. \cite{peller-konrad_memory_2023}. Their system has a working memory, a \ac{STM}, and \ac{LTM}. The memory mediates between high-level components processing symbolic data and the lower level dealing with sensor data.

Verma et al. review papers that are pursuing a hippocampus-inspired approach for cognitive architectures \cite{verma_akshat_pdf_2024}. They find that there are numerous approaches that accurately simulate parts of the hippocampus, which is responsible for transferring memories from the \ac{STM} into the \ac{LTM}.
Jockel et al. propose a checklist to evaluate an artificial episodic memory with \cite{jockel_towards_2008}.
Both papers conclude that more interdisciplinary work is needed to close the gap between biological and computational models \cite{verma_akshat_pdf_2024,jockel_towards_2008}.

\subsection{Architectures to Enable Natural Language in ITL}
To incorporate natural language into a robotic object learning and manipulation scenario, the \textit{Pamini} dialog system was developed \cite{peltason_pamini_2010,peltason2011curious}. It aims to disclose the system's internal state in a natural dialogue, and makes executions interruptible and configurable online. \textit{Pamini} was combined with \textit{InproTK} \cite{baumann_inprotk_2012} to allow for incremental feedback and dynamic task executions \cite{carlmeyer_towards_2014}.
To understand the meaning of words in the context of their sentence, Schlangen et al. \cite{schlangen_general_2009} proposed a system to assess sentences in incremental units.
Thanks to \acp{LLM}, recent approaches have little problems with a lack of the linguistic coverage or context-induced ambiguity of sentences. One such approach is \textit{SPIRES} that conceptualizes natural language by using \acp{LLM} and ontologies \cite{caufield_structured_2024}.
Also, recent architectures focus on the use of \acp{LLM} for the high-level orchestration of robot abilities and learning from mistakes~\cite{baermann2024}.

\subsection{Multi-modality in Architectures for ITL}
Steil et al. present the robotic learning architecture \textit{GRAVIS}, which uses multi-modal input for grasping tasks~\cite{steil2004situated}.
Grounded in the approach of Schlangen et al., Kennington et al. propose a network architecture incorporating multi-modality \cite{schlangen_general_2009,kennington_incremental_2021}.
However, both architectures lack memories.
Wang et al. also build a network to reason on multiple modalities but use a transformer model to join visual and speech information \cite{wang_multi-modal_2024}. A symbolic approach is followed by the \textit{EMISSOR} framework that stores multi-modal episodic memories in the human-readable format \cite{baez_santamaria_emissor_2021}.

\subsection{Knowledge representation for task learning}
\textit{Knowrob} is presented by Beetz et al. and is capable of symbolic learning on multi-modal \acp{NEEM}. It employs various ways of learning and reasoning to generate action plans for robots \cite{beetz_knowrob_2018,beetz_cram_2010}. KnowRob supports multiple action selection strategies and grounds its reasoning in ontologies \cite{gangemi_sweetening_2002, masolo_wonderweb_2003, besler_foundations_2021}.
A hybrid-approach is used by Akinobu et al. who reinforce an episodic memory with an \ac{LLM}  to make a robot find household items \cite{akinobu_mizutani_hippocampus-inspired_2024}. It utilizes a common and an environment-specific knowledge base. The latter represents the hippocampus, storing episodic memories. When there is no memory of the location, an \ac{LLM} is queried for common knowledge of where to search.

To sum up the related work, several cognitive architectures from different research areas have been implemented. To enable natural language interaction for robotic task learning, the cognitive architecture must meet several characteristics, which we highlight below:

\begin{enumerate}
    \item Support Natural Language input.
    \item Support synchronized input of multiple modalities.
    \item Assess and store the natural language input from different viewpoints/ tokenizations.
    \item Timestamp the natural language to interpret it in its environmental context.
    \item Have a basic semantic understanding of objects and actions for the scenario at hand.
\end{enumerate}

\section{Cognitive Architectures for \ac{CCTL}}
\label{sec:concept}

The primary challenge in robotics for everyday activities lies in the unstructured nature of both the environment and the tasks to be performed. It is not feasible to deploy robots with all the required knowledge. Also, due to the small nuances that make everyday tasks highly complex, prompting or few-shot learning may not convey all the required details to successfully execute them. Instead, a robot should be designed to interact with humans and acquire needed skills through dynamic learning, similarly to children. Given that the human cognition is highly efficient and adaptable, we propose that a cognitive architecture can enable robots to achieve comparable adaptability and learning capabilities.

We first examine what characteristics of cognitive architectures are obligatory for \ac{CCTL}. Therefore, we use the proposed feature sets based on the work of Kotseruba et al., Laird et al., and Sun \cite{kotseruba_40_2020,laird_cognitive_2010,sun_desiderata_2004}. Then we assess how \ac{CCTL} benefits from each feature.
Secondly, we introduce a concept that tightly links concepts from the realm of educational science, neuroscience and robotics. The concept emphasizes the specific features of cognitive architectures and their roles in enabling \ac{CCTL}.

\subsection{Requirements for Natural Language in \ac{CCTL}}
\label{sec:concept:characteristics}
As already mentioned in section \ref{sec:related_work}, Sun defined four core requirements for cognitive architectures, namely ecological, bio-evolutionary and cognitive realism and lastly eclecticism of methodologies and techniques \cite{sun_desiderata_2004}. On a lower level, Kotseruba et al. assessed a multitude of different cognitive architectures by the categories perception, attention mechanisms, action selection, learning, memory, reasoning, and meta-reasoning \cite{kotseruba_40_2020}. Going into detail, Laird et al. defined a set of characteristics needed for cognitive architectures and their environments \cite{laird_cognitive_2010}.

\begin{figure*}
    \centering
    \includegraphics[width=1\linewidth]{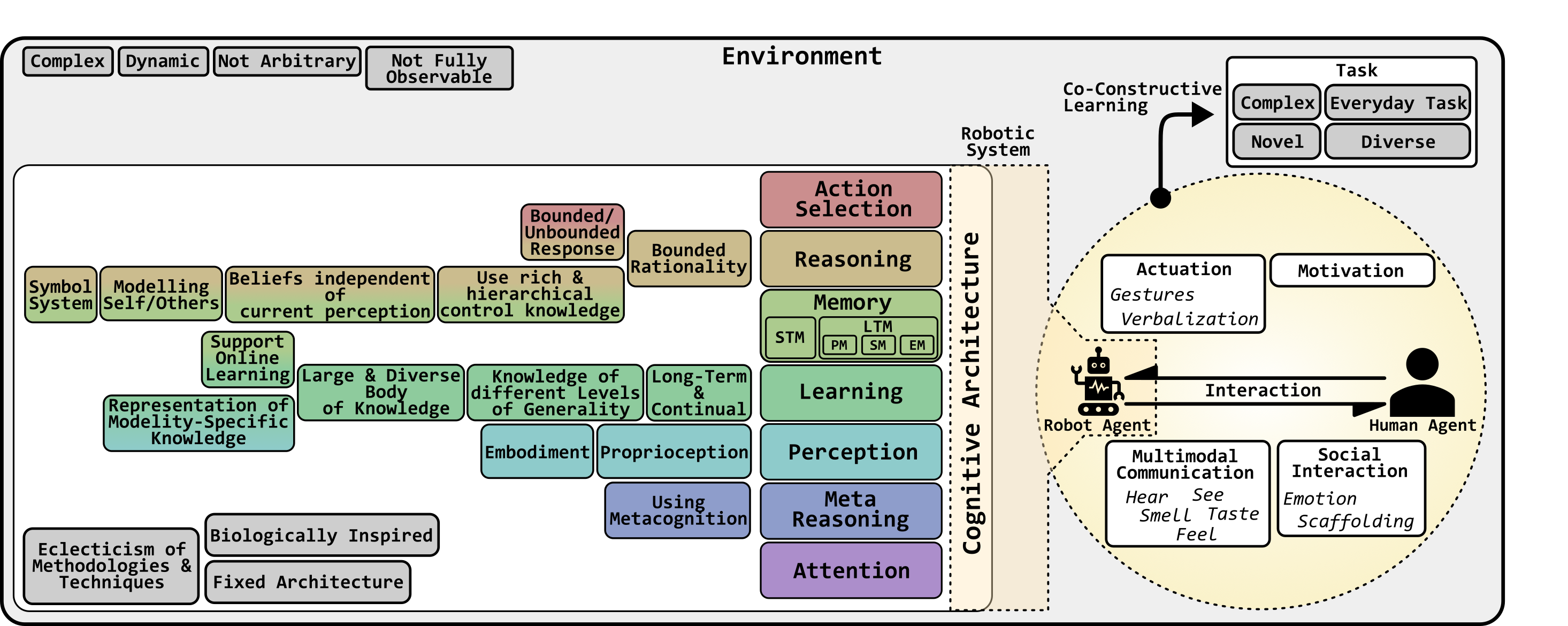}
    \caption{The cognitive architecture must have the right features to support an interactive dialogue between the human and the robot agent. The Memory Component is divided into \ac{STM} and \ac{LTM}. The \ac{LTM} is further divided into its three sub-systems Procedural Memory (PM), Semantic Memory (SM) and Episodic Memory (EM).}
    \label{fig:cog_arc_order}
\end{figure*}

Building on the collective work of these three sources, we developed a set of requirements, which are presented in Figure \ref{fig:cog_arc_order}. We state that all of the shown characteristics of a cognitive architecture are relevant to \ac{CCTL} whilst the extent to which a feature is incorporated and also the set of supported modalities and actuation naturally varies across robotic systems. We assert that \ac{CCTL} is essentially shaped by the capabilities of the cognitive architecture in these categories.

In the analysis, we focus our attention on \ac{HRI}, but many of the aspects are also applicable to scenarios between two robot agents.
The following list puts each feature from Figure \ref{fig:cog_arc_order} in the perspective of \ac{CCTL} and explains its relevance.

\begin{itemize}
    \item \textbf{Perception} For any interaction to take place, there must be a way for the system to receive input from the environment. The natural language is a powerful modality to precisely convey information and to add details, as further outlined in section \ref{sec:background:natural_language}.

    \item \textbf{Attention} There is plenty of information in every piece of interaction. Especially in a learning scenario, a lot of novel information is being conveyed by different modalities. The human cognition filters out most of the information and puts its attention onto the bits of interaction that are meaningful to create knowledge. A robot without an attention mechanism will be overwhelmed by the amount of data that is rushing in. It surely is imaginable that a sufficiently powerful system can be employed to record all the information. But learning from this data is unfeasible, due to a low signal-to-noise ratio. There must be a mechanism that selects what information is valuable and worth keeping.

    \item \textbf{Reasoning} It is a natural requirement of any interaction to be reasonable. That is, humans take reasoning in the interaction partner for granted. Thus, \ac{CCTL} in \ac{HRI} can only be meaningful when the robot makes reasoned decisions during the interaction. Unreasonable behavior would alienate the robot, and the principles that constitute the frame of \ac{CCTL} would break.
    
    \item \textbf{Meta Reasoning} In the interaction, the robotic system needs to know when it misses knowledge to consequently initiate a learning interaction. Therefore, it needs meta information about its capabilities and expertise and the lack of knowledge.
    
    \item \textbf{Action Selection} An interaction is only considered an interaction when both parties are taking part in it (See Section \ref{sec:background:interactive_task_learning}). In the context of a learning scenario, it is crucial for both agents that the other party is providing feedback. The teacher needs to know when the learning strategy is beneficial to the student, whilst the student needs feedback to assess if the novel knowledge is understood correctly. Also, feedback from the teacher is needed to understand when an entirely new task is being explained or when additional information to a known task is given.

    \item \textbf{Memory} The memory system must be able to store necessary semantic information for the context at hand and also to memorize novel concepts. It must be able to put the information of one piece of interaction into the context of a \textit{Pragmatic Frame} as a whole. Also, the memory is mandatory to keep knowledge and to transfer it onto novel situations.
    
    \item \textbf{Learning} When the system does not know a concept that occurs within the interaction, it must be able to learn it. Interaction is built around the concept of starting off without a common ground of understanding and then mutually gravitating during the learning interaction. Therefore, the robot must learn and adapt online. It is not sufficient to reflect on the contents of the interaction in retrospection.
\end{itemize}

\subsection{The importance of Pragmatic Frames for Co-Construction}
\label{sec:concept:pragmatic_frames_and_co-construction}
\ac{CCTL} can only occur in interactions where the robot is capable of engaging meaningfully. Unlike \ac{ITL}, which primarily focuses on the robot's ability to learn from direct human instructions or feedback, Co-Construction involves a deeper, collaborative process and the creation of a mutual understanding during interaction, in which both parties optimize their communication implicitly to enhance the learning process. This process is extensively described in \cite{vollmer_interactive_2023}.

\begin{figure}
    \centering
    \includegraphics[width=1\linewidth]{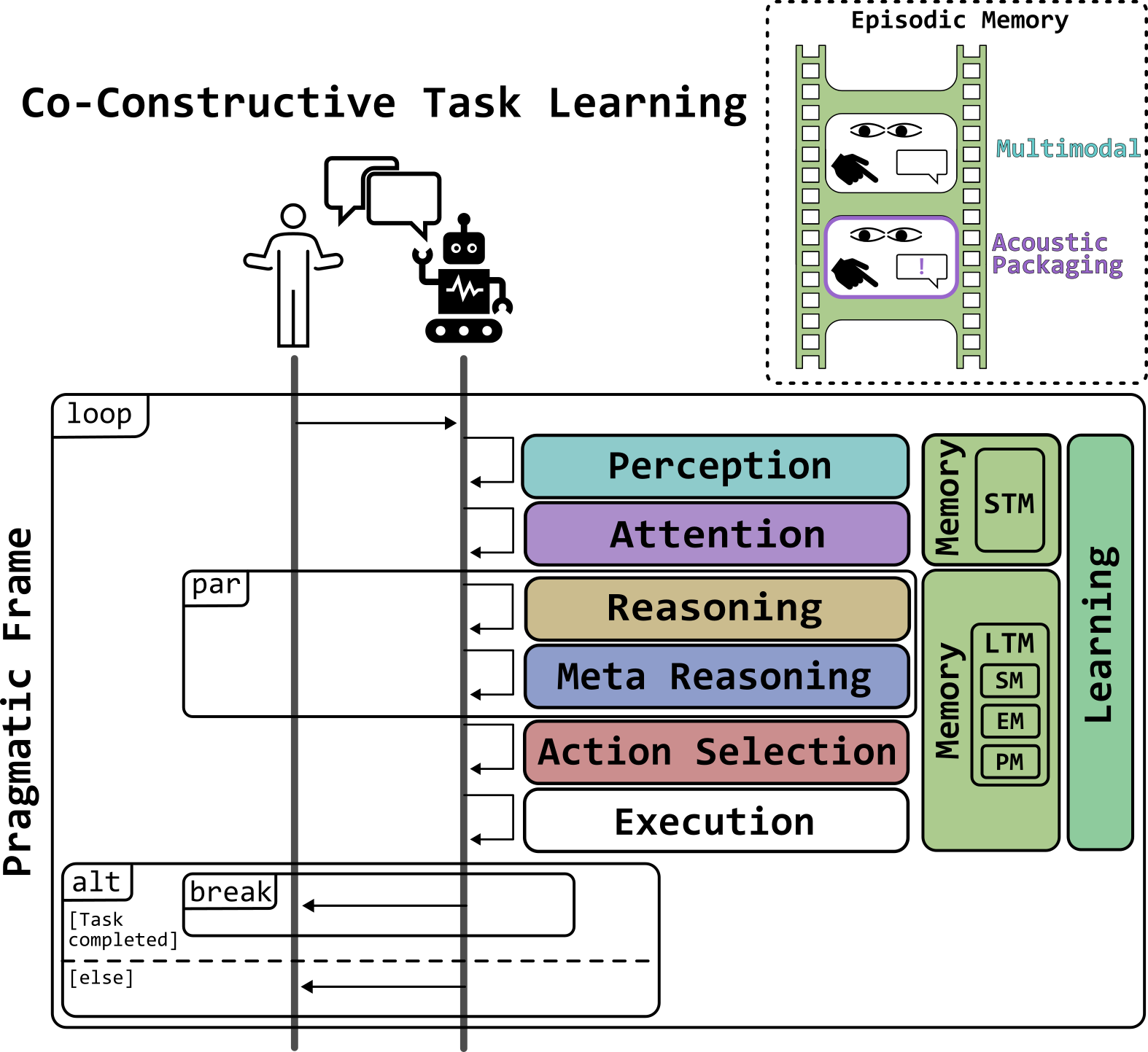}
    \caption{The different components of a cognitive architecture are all demanded during \ac{CCTL} and span a \textit{Pragmatic Frame}. The visualization for the looped process (\texttt{loop}), parallel executions (\texttt{par}), exit condition (\texttt{break}) and conditional executions (\texttt{alt}) is inspired by \ac{UML} sequence diagrams.}
    \label{fig:mem_sys_timeline}
\end{figure}

Figure \ref{fig:mem_sys_timeline} visualizes an abstract scenario that show how all components of a cognitive architecture come into play during \ac{CCTL}. The process is initiated by the human agent, who verbally orders the robot to execute some task that might not be fully known to it. 
The system recognizes the human input thanks to its verbal perception capabilities and stores it intermediately in the \ac{STM} to tokenize it. Thereby, the input can be further analyzed from different viewpoints. Using attention mechanisms, e.g. \textit{Acoustic Packaging} for the verbal input, the cognitive architecture sorts out irrelevant input to focus on the important clues and information in the reasoning process.

Reasoning and meta reasoning can both run in parallel. Using the sub-systems of the \ac{LTM}, the robot assesses what the task request is and what knowledge is missing to successfully execute it. Based on the results of the reasoning process, the robot selects an appropriate action. This can for example be to interact with the human agent to obtain additional cues about the task or to request a demonstration. The result of the reasoning might also be that the information is sufficient for executing the task, which could lead the robot to output an affirmation and begin acting.

Based on the prior selection process, the robot executes some routine that manifests in an interaction with the human agent. The robot might utter a question now, expresses how its current understanding of the task is, moves some actuator to do a pointing action or might even start to execute the demanded task. In doing so, it actively shapes the co-construction that is unfolding and thereby engaging the human teacher to respond in a valuable manner.
When it verbalizes something, it might itself use scaffolding mechanisms, e.g. hesitation or changing the pitch to direct the human's attention.
The procedure can be understood as a learning process for the robot. Based on the expressed question, the answer of the user might generate novel knowledge that enriches the semantic memory, a task demonstration can extend the procedural memory by new procedures or fine-tune parameters and lastly, the entirety of the interaction with the human agent can be stored in an episodic memory.
The episodic memory that is being stored is again harnessing the capabilities of the cognitive architecture's components as it stores multi-modal information depending on the perceptive capabilities of the robot and uses the results of the attention mechanism to mark important information.

The process of perceiving the human agent's input, reasoning on it, selecting a responsive action and execution is looping until the task is successfully executed (See Figure \ref{fig:mem_sys_timeline}). The completion of the task is either recognized by the robot itself or indicated by the human agent.
Each iteration creates new episodic memories and might as well result in changes to the procedural and semantic memory. Moreover, the online communication channel enables the teacher to immediately interfere with the robot's behavior in case it misinterpreted the given input or the perceived intentions.

The process from the first query of the human agent, the following sequential actions, up to the achievement of a joint goal is considered a \textit{Pragmatic Frame} \cite{rohlfing_alternative_2016}. Within this frame, both agents select their actions with the other agent dynamically during the course of the interaction with the objective to optimize the learning outcome by employing modalities of communication that the interaction benefits most from.

\subsection{Towards Implementation Architecture}

\label{sec:concept:towards_implementation_architecture}
We are working towards a cognitive architecture that is capable of supporting \ac{CCTL}, as depicted in Figure \ref{fig:mem_sys_timeline}. Big steps have already been achieved in implementing architectures that support all the obligatory characteristics, as summarized in section \ref{sec:related_work}. Many of the aspects that were summed up in Figure\ref{fig:cog_arc_order} are covered by capable cognitive architectures e.g. \textit{KnowRob} \cite{beetz_knowrob_2018,beetz_cram_2010}. Moreover, state-of-the-art memory systems come with the mandatory sub-systems to support the necessary mechanisms that are needed in a \ac{CCTL} scenario \cite{peller-konrad_memory_2023,verma_akshat_pdf_2024,vinanzi_casper_2024}.
However, we see some gaps hindering such architectures to be utilized for human-robot interaction.
One of the downsides of current systems is that the process of recording new episodic memories must be manually orchestrated, although there are tools e.g. \textit{RobCogBridge} to simplify the recording by automatization \cite{manuel_scheibl_manuel_2024}.

The second and more important gap we see lies in the supported modalities, as currently mostly video and tactile input are supported. That means, the user can either demonstrate to the robot how to conduct some task or it can manipulate the robot's position to navigate it towards the goal.
Natural language, especially in a multi-modal scenario, is rarely considered, although it might be the most important modality in \ac{HHI} as elaborated in section \ref{sec:background:natural_language}. To engage in \ac{CCTL}, the human must not only be able to give the initial command to the robot in natural language, but to give additional information e.g. clarifications or details in the ongoing interaction with the robot. We also want to emphasize that due to attention mechanisms like \textit{Acoustic Packaging} (see Section \ref{sec:background:natural_language}) a textual interface does not suffice as it would limit the synchronicity and thus the leveraging effect of multi-modality.

Lastly, we believe that the concept of \textit{Pragmatic Frames} (see section \ref{sec:background:interactive_task_learning}) needs to find its way into \ac{HRI} to enable \ac{CCTL} as described and depicted in Section \ref{sec:concept:pragmatic_frames_and_co-construction}.
Without a frame in which the robot can reason about the interaction as a whole, it will not be able to meaningfully shape the interaction.
The \textit{Pragmatic Frame} in \ac{CCTL} enables the system to correlate the type of information provided by the human with its own actions and behavioral patterns. Thereby, it enables the robot to more efficiently close the lack of knowledge by reasoning, e.g., what question to ask or what modality to use next.

To close these gaps, we see several points in need of integration into the set of capabilities of a cognitive architecture, e.g. \textit{KnowRob} for \ac{CCTL} scenarios:
\begin{itemize}
    \item Integration of a mechanism to recognize the beginning of a \ac{CCTL} scenario. Therefore, the cognitive architecture must have \textit{Meta Reasoning} capabilities to quickly recognize and assess its lack of knowledge.
    \item Integration of a mechanism to recognize the end of a \ac{CCTL} scenario. We added this as a disjoint item from the first one since we want to emphasize that only recognizing both, the start and end, suffices to border the \textit{Pragmatic Frame} that unfolds.
    \item Integration of natural language. For an \ac{HRI} scenario, it is obligatory that the human can interact with the robot as they would with another human in \ac{HHI}. Only when the interaction is natural, the intrinsic benefits of a \textit{Pragmatic Frame} as it takes place in \ac{HHI} will improve the learning performance in \ac{HRI}.
    \item Usage of the various layers of natural language. As pointed out in section \ref{sec:background:natural_language}, there is much more to spoken language than words. Humans convey many layers of information that the robot can harness. Therefore, scaffolding strategies like hesitation in the speech and the pitch of the voice should be recognized and memorized.
    \item Leveraging episodic memories in the learning process. The episodic memory is of outstanding importance for a human to learn novel concepts and tasks (see section \ref{sec:background:memory}). Many cognitive architectures have implemented an episodic memory, but many architectures do not utilize it to extract new knowledge.
    \item True synchronicity of multi-modality. Attention mechanisms like \textit{Acoustic Packaging} that have been introduced in section \ref{sec:background:natural_language} can only be used when the different modalities that highlight important pieces of information among each other can be recorded and analyzed simultaneously. Most cognitive architectures do not connect the data from multiple modalities to recognize the important bits.
\end{itemize}

\section{Conclusion}
\label{sec:conclusion}
In this paper, we identified the key characteristics a cognitive architecture must possess to enable natural language interaction in task learning.
We integrated different concepts from various disciplines and merged these concepts to establish a foundation for \acl{CCTL}.
Furthermore,  we presented a concrete combination of implemented concepts and highlighted the existing gaps that must be addressed to meet the requirements.
We conclude that there are plenty of implementations that fulfill the requirements to be considered a cognitive architecture, but that they do not yet meet the level of complexity to hold well in a \ac{CCTL} scenario. In particular, we emphasize the shortcomings in natural language capabilities. Finally, we pointed out a concrete set of missing requirements in the \textit{KnowRob} architecture that must be addressed for it to be suitable for \ac{CCTL} in \ac{HRI}.


\bibliographystyle{plain}
\bibliography{bibliography}

\end{document}